\documentclass{article}
\usepackage{spconf,amsmath,graphicx}
\usepackage{amsfonts} 
\usepackage{algorithm} 
\usepackage{algpseudocode} 
\usepackage{xcolor}
\usepackage{subcaption}
\usepackage{mwe}
\usepackage{adjustbox}

\title{Clustering Plotted Data by Image Segmentation}
\threeauthorss
 {Tarek Naous}
	{American University of Beirut}
 {Srinjay Sarkar}
	{VinAI Research}
 {Abubakar Abid, James Zou}
	{Stanford University}

\begin{document}
%
\maketitle
\begin{abstract}
Clustering algorithms are one of the main analytical methods to detect patterns in unlabeled data. Existing clustering methods typically treat samples in a dataset as points in a metric space and compute distances to group together similar points. In this paper, we present a wholly different way of clustering points in 2-dimensional space, inspired by how humans cluster data: by training neural networks to perform instance segmentation on plotted data. Our approach, \textit{Visual Clustering}, has several advantages over traditional clustering algorithms: it is much faster than most existing clustering algorithms  (making it suitable for very large  datasets), it agrees strongly with human intuition for clusters, and it is by default hyperparameter free (although additional steps with hyperparameters can be introduced for more control of the algorithm). We describe the method and compare it to ten other clustering methods on synthetic data to illustrate its advantages and disadvantages. We then demonstrate how our approach can be extended to higher dimensional data and illustrate its performance on real-world data. The implementation of Visual Clustering is publicly available and can be applied to any dataset in a few lines of code.

\end{abstract}
\section{Introduction}
\label{sec:intro}
Numerous applications require the classification of unlabeled samples in dataset into disjointed clusters such that samples within the same cluster are similar yet samples in different clusters differ meaningfully. Many such clustering algorithms have been developed satisfying different desiderata for applications in  fields such as image processing \cite{T1}, biomedicine \cite{article}, and spatial data \cite{ester1996density}.  

The most commonly used clustering algorithms, such as K-means clustering \cite{Jin2017} , Gaussian mixture clustering \cite{Jin2010}, and  DBSCAN \cite{dbscan}, treat samples as points in a metric (usually Euclidean) space and group together points based on distances to other points or to computed exemplars. For example, the K-means clustering identifies optimal centroids in the metric space to which the distance of all samples in the dataset are minimized. The Gaussian mixture algorithm assumes the data is sampled from a mixture of Gaussians and produces clusters in the data to maximize the likelihood, which happens when data points are close to the centers of the Gaussian distributions. DBSCAN is a density-based clustering algorithm which does not assume the number of clusters for the given dataset, but considers a group of point belonging to the same cluster if there are a certain number of points in the neighborhood the selected point. The clusters are expanded by recursively considering distances to all other points.

Since most of these algorithms involve measuring distances between points, they scale poorly to large datasets with millions or billions of samples. In our work, we introduce a completely different kind of clustering algorithm, designed for two-dimensional datasets for large datasets. Our approach, which we call \textit{visual clustering}, is inspired by how humans cluster data: rather than computing distances, we segment data points into clusters based on the shape of large regions within the dataset. We simulate this process by training neural networks to perform instance segmentation on plotted data. Our approach has several advantages over traditional clustering algorithms: (1) Because the main step in the algorithm is running a prediction from one neural network, it is much faster than most existing clustering algorithms and scales easily to datasets with millions or billions of samples. (2) As we show on many kinds of datasets, it agrees strongly with human intuition for clusters, moreso than many other clustering algorithms. (3) The core algorithm is hyperparameter free, although we suggest additional steps with hyperparameters that can be introduced for more control of the algorithm. Our implementation is publicly available \footnote{https://github.com/tareknaous/visual-clustering}.

Clustering has been applied in the literature to solve various problems in deep learning and computer vision such as unsupervised image segmentation \cite{R1} facial landmark detection \cite{R6}, and image grouping \cite{R2,R7}. However, no previous work has leveraged the fast inference time of trained neural network models to perform clustering. Developing deep learning models that can replace classical algorithms has been studied in the literature for various problems such as sorting \cite{nn-sort}, solving mixed integer problems \cite{R4}, or even replacing index structures in data management systems \cite{R5}. To the best of our knowledge, our work is the first to introduce an algorithm based on a supervised deep learning model to perform clustering  of numerical datasets.

\begin{figure*}[h]
    \centering
    \includegraphics[width=0.9\linewidth]{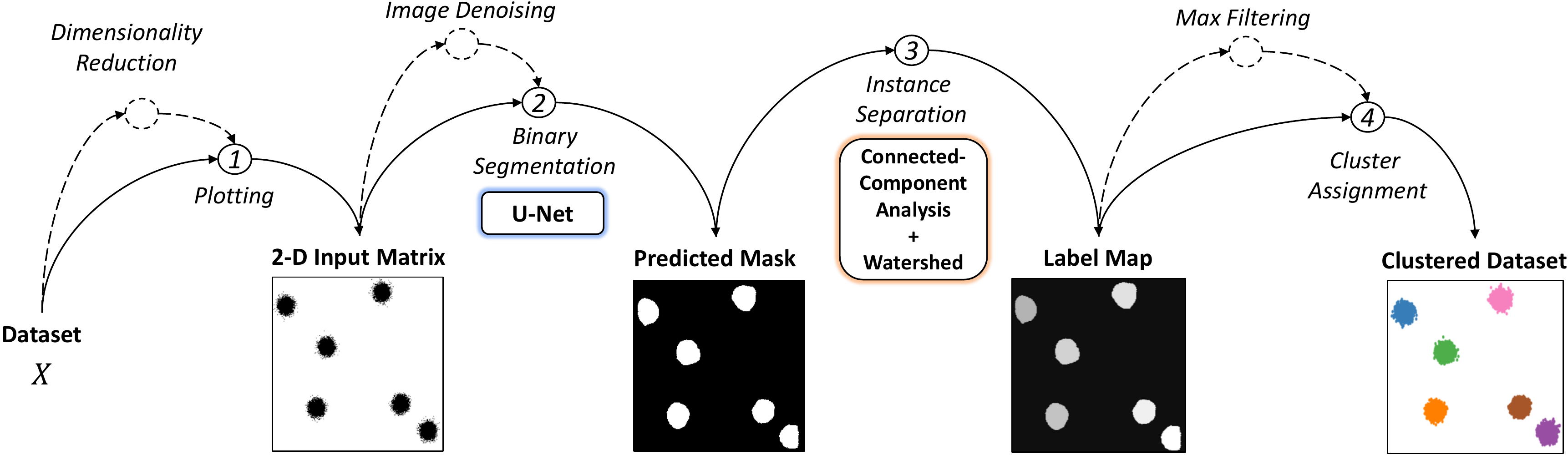}
    \caption{Diagrammatic view of our Visual Clustering algorithm. Dotted lines infer optional steps. The algorithm first creates a matrix representation of the dataset that is used as input to a binary segmentation model (U-Net). Connected component analysis and watershed are applied to the predicted binary mask to separate the different instances in the image, resulting in a label map from which cluster assignment is finally performed.}
    \label{fig:pipeline}
\end{figure*}

\section{Methods}
\label{sec:methods}

\subsection{Core Algorithm}

Our proposed Visual Clustering algorithm is illustrated in Figure~\ref{fig:pipeline} and consists of four main steps that we describe in this section. Consider the dataset $X \in \mathbb{R}^{m \times 2}$ that we would like to cluster. We start by the plotting step where $X$ is represented in a two-dimensional matrix form, denoted by $I(X)$. This is done by first linearly shifting the values of both features to be $\in [0,256]$ and then filling a zero-initialized $256\times256$ matrix by a value of 1 for each sample in the dataset according to its coordinates. In case the dataset was high-dimensional, we apply Principal Component Analysis (PCA) and use the first two principle components as features. The matrix, which can be visualized as an image, is then fed as an input to the second step of binary segmentation where a pre-trained binary segmentation model is used. We adopt the U-Net architecture \cite{unet} for binary segmentation and train it in supervised fashion on images of plotted datasets, which were synthetically generated, and their binary masks. The U-Net model predicts a pixel-level binary mask $\hat{M}(I(X))$.

The predicted binary mask by the trained U-Net model contains information on where cluster areas are located. However, the binary mask alone does not indicate how many clusters there are in the image. Hence, the next step in our approach is separating the instances (or clusters) present in the binary mask. To do this, we apply Connected Component Analysis on the predicted mask followed by a Watershed transformation for instance separation. This results in a label map $L$ where pixels belonging to the same cluster are assigned the same label value. The final step in our approach is cluster assignment, where we assign a cluster label to each sample in the dataset based on its location in the label map.

\subsection{Training the Binary Segmentation Model}
To train the U-Net model for binary segmentation, we generated 1,000 synthetic datasets of blob-shaped clusters. To create the label for each dataset, we computed the convex hull of each cluster in the dataset. The convex hulls are then used to form a the binary mask label. When the hulls of two clusters were intersecting beyond a threshold of 30\%, they were joined together to form one cluster. If the intersection was below the threshold, the hulls were subtracted to separate the clusters in the binary mask. The U-Net model achieved a test-set Intersection-Over-Union (IOU) of 88.7\%.

\begin{figure*}[h]
    \centering
    \includegraphics[width=0.65\linewidth]{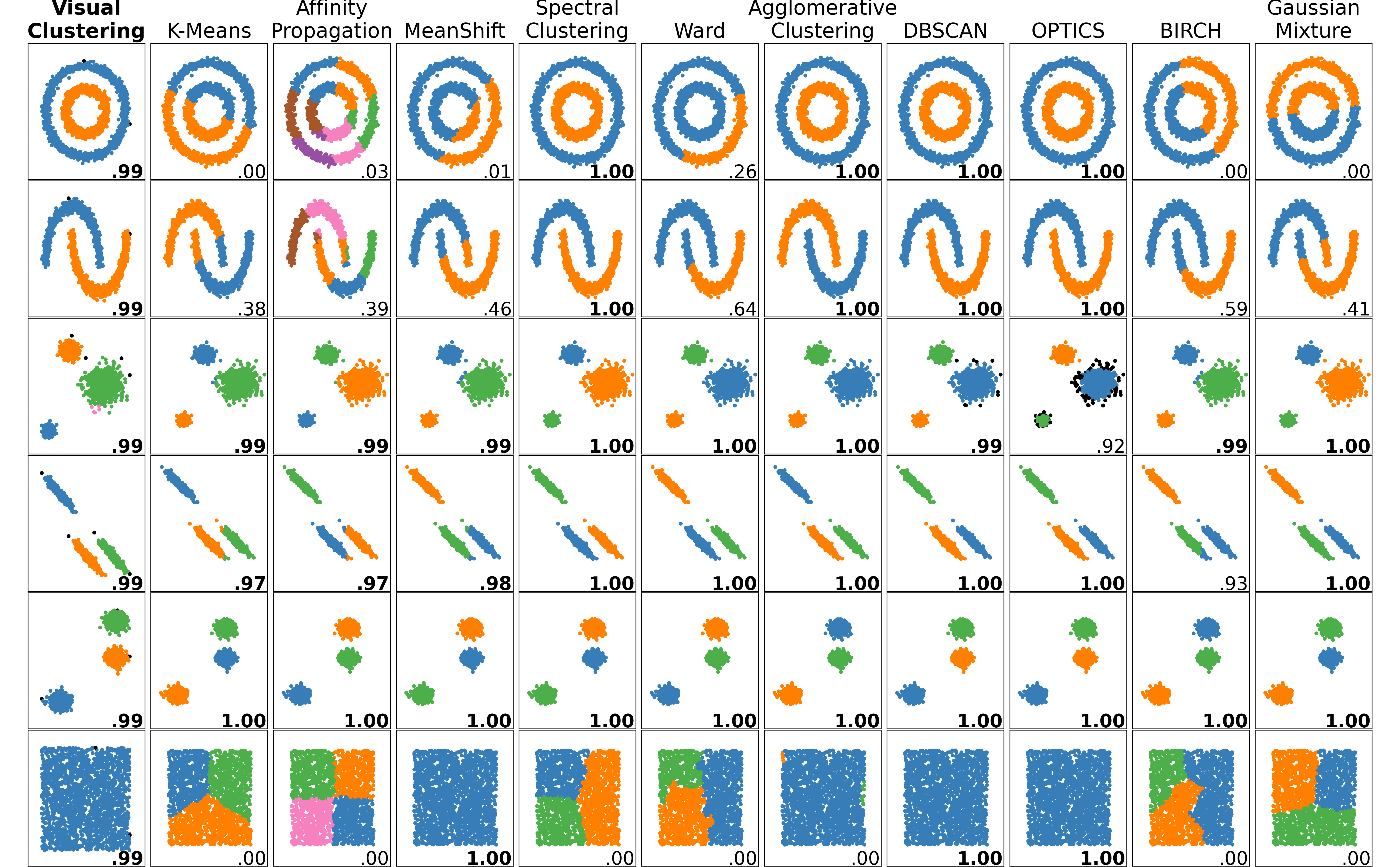}
    \caption{Comparison of Visual Clustering with classical clustering algorithms on synthetic datasets with several cluster shapes. Bottom right numbers indicate the Adjusted Mutual Information Score between the ground truth labeling and the predicted labeling. Scores that are above 0.95 are highlighted in bold. Visual Clustering and DBSCAN are the only two algorithms that achieves nearly perfect matches with ground-truth labeling on all synthetic datasets.}
    \label{fig:clustering-comparison}
\end{figure*}

\subsection{Handling Inseparable Clusters with Image Denoising}

The developed binary segmentation model segments the image in a way that identifies disconnected clusters. If two or more clusters intersect, the model is likely to combine these clusters into one. This holds true especially in real world datasets where clusters are more likely to be inseparable. To avoid this problem, we introduce an optional pre-processing step of image denoising to filter out low density areas in the image and emphasize high density areas. This helps disconnect cluster regions that seem to be connected by sparse data points. We specifically denoise the image through a median filter, where increasing the size of the filter will eliminate low density areas and emphasize high density ones more severely.

\subsection{Handling Unassigned Points with Max Filtering}
When assigning labels to each point in the dataset in the final step of the algorithm, many points will fall in regions that are outside but near the cluster area. These may be points that deviate from where the majority of the points in the cluster are located and thus would be ignored by the binary segmentation model. Unassigned points may also be low density areas that were filtered out in the image denoising step, if used. To address this, we perform an optional maximum filtering operation on the label map to increase the area of each cluster, helping cover nearby unassigned points. The degree in which we would like to increase the cluster areas is controlled by the size of the max filter, where a larger filter size will result in larger cluster areas.

\section{Results}
\label{sec:results}


\subsection{Clustering Performance on Synthetic Datasets and Computation Time Comparison}

The performance of our visual clustering approach is compared to multiple classical clustering algorithms on synthetic datasets of various cluster shapes in Figure~\ref{fig:clustering-comparison}. Although the segmentation model used in our algorithm was only trained on blob-shaped clusters, it is able to successfully segment clusters independently of their shape. On datasets that have more complex patterns, such as circle or moon-shaped clusters, our algorithm provides clustering results that are more in-line with human intuition compared to the results of K-means, Affinity Propagation, or Gaussian Mixture. While other algorithms such as DBSCAN or Spectral Clustering agree with human intuition, they suffer from a large computation time and cannot be used to cluster large datasets efficiently.


\begin{table}[h]
\centering
\resizebox{0.49\textwidth}{!}{%
\begin{tabular}{lcccccc}
\cline{2-7}
\multicolumn{1}{c}{}                   & \multicolumn{6}{c}{\textbf{Number of Samples}}                \\ \hline
\multicolumn{1}{c}{\textbf{Algorithm}} & 10K    & 50K      & 100K     & 500K     & 1M       & 2M       \\ \hline
\textit{\textbf{Visual Clustering}}    & 0.292  & 0.571    & 0.909    & 3.686    & 7.222    & 14.096   \\
K-Means                                & 0.155  & 0.541    & 1.103    & 5.470    & 9.519    & 18.956   \\
Affinity Propagation                   & 175.35 & $\infty$ & $\infty$ & $\infty$ & $\infty$ & $\infty$ \\
MeanShift                              & 3.482  & 101.82   & $\infty$ & $\infty$ & $\infty$ & $\infty$ \\
Spectral Clustering                    & 0.052  & 0.509    & 0.796    & 7.455    & 20.001   & 53.559   \\
Ward                                   & 1.994  & 27.965   & 93.564   & $\infty$ & $\infty$ & $\infty$ \\
Agglomerative Clustering               & 1.177  & 12.154   & 39.886   & $\infty$ & $\infty$ & $\infty$ \\
DBSCAN                                 & 0.093  & 0.357    & 0.837    & 7.604    & 20.009   & 52.648   \\
Optics                                 & 16.515 & $\infty$ & $\infty$ & $\infty$ & $\infty$ & $\infty$ \\
BIRCH                                  & 1.298  & 7.390    & 14.320   & $\infty$ & $\infty$ & $\infty$ \\
Gaussian Mixture                       & 0.089  & 0.358    & 0.726    & 3.047    & 5.949    & 11.962   \\ \hline
\end{tabular}
}

\caption{End-to-end computation time (in seconds) comparison of Visual Clustering with classical clustering algorithms for an increasing number of samples. $\infty$ indicates a computation time beyond 3 minutes and hence was not included. Visual Clustering is as fast as Gaussian Mixture and faster than all the other classical algorithms on large datasets.}
\label{tab:computation-time}
\end{table}

\begin{figure*}[h]
    \centering
    \includegraphics[width=0.7\linewidth]{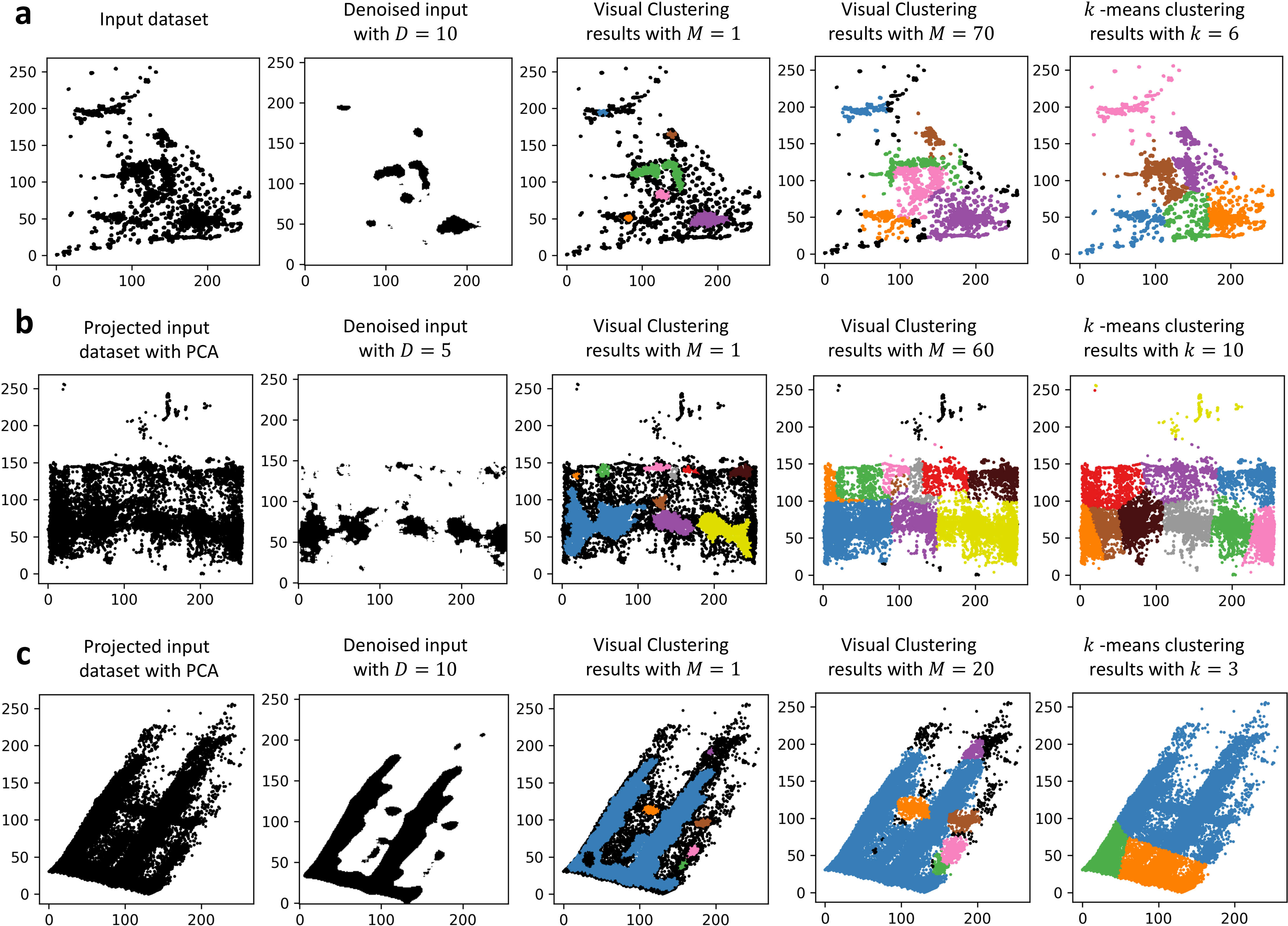}
    \caption{Results of the Visual Clustering algorithm on three real-world datasets  compared with $k$-means clustering. $D$ stands for the kernel size of the median filter. $M$ stands for the kernel size of the max filter. Dataset \textbf{(a)} consists of urban road accidents coordinates. It contains 2 features and 360,177 samples. Dataset \textbf{(b)} consists of geo-magnetic field data for indoor localisation. It contains 13 features and 58,374 samples. Dataset \textbf{(c)} consists of individual household electric power consumption data. It contains 7 features and 2,075,259 samples. Both datasets \textbf{(b)} and \textbf{(c)} are reduced to two dimensions using PCA. We note that we cannot conclude which algorithm provides the best clustering since no ground truth labels are available in these datasets.}
    \label{fig:results-real-datasets}
\end{figure*}


The computation time of the segmentation model, instance separation algorithms, and filtering techniques are independent of the number of samples in the dataset. As our algorithm is a completely vision-based approach, it provides a great computation time advantage. However, the computation time of the plotting and cluster assignment steps in our algorithm increase linearly with the number of samples.  In Table~\ref{tab:computation-time}, we show an end-to-end computation time comparison between visual clustering (including the plotting time) and  classical algorithms for an increasing number of samples. Visual Clustering achieves a very fast computation time that is almost identical to Gaussian Mixture, which was the fastest classical algorithm among the ones tested, and faster performance than K-Means clustering. Visual Clustering also outperforms all the rest of the classical algorithms in terms of computation time. Therefore, Visual Clustering achieves a compromise between slow classical algorithms like Affinity Propagation that can cluster complex patterns in a human intuitive manner while having a very fast computation time like Gaussian Mixture or K-Means.

\subsection{Performance on Real-world Datasets}

We evaluate Visual Clustering on three real-world datasets obtained from the UCI repository among which one is two-dimensional and the two others are of higher dimension. The clustering results on those datasets are shown in Figure~\ref{fig:results-real-datasets}. In real world datasets, clusters are more likely to not be visually separable. For the first two datasets (\textbf{a} and \textbf{b}), the image denoising step shows its effectiveness in emphasizing high density regions in the dataset and eliminating low density regions. This helps the binary segmentation model capture more clusters as they become visually separable. The majority of the points that belong to low density regions were then recaptured by performing max filtering on the label map prior to cluster assignment.

The third dataset (dataset \textbf{c}) presents a more challenging case for Visual Clustering where most of the points in the plot are connected through the same pattern. While visually it would be intuitive to identify the three main lines shown more clearly after denoising as three different clusters, the algorithm instead considers them as one cluster. This is because Visual Clustering relies on binary segmentation and connect component analysis which makes it difficult to identify several clusters on connected patterns, which could or could not be desirable output based on the domain expertise of users. In this respect, our future work will focus on improving the Visual Clustering algorithm to enable further flexibility in segmentation in a way that provides an ability in placing multiple clusters on connected patterns. We will also investigate the problem of assigning clusters to sparse points in the dataset which do not get covered even after performing max filtering on the label map.

\section{Conclusion}
\label{sec:conclusion}

We introduced Visual Clustering, a fast clustering algorithm based on a trained image segmentation model. Visual Clustering is inspired by how humans cluster data: by plotting datasets in 2D and identifying groups of similar points. Our experiments on real and synthetic datasets and comparisons to ten classical clustering algorithms show that Visual Clustering achieves clustering results that are in-line with human intuition in a fast computation time that outperforms almost all the rest of the classical algorithms, making it efficient for usage on very large datasets.

\bibliographystyle{IEEEbib}
\bibliography{refs}

\end{document}